# Signature Region of Interest using Auto cropping

Bassam Al-Mahadeen[1], Mokhled S. AlTarawneh[2] and Islam H. AlTarawneh[2]

[1] Math. And Computer Department, Faculty of science, Tafila Technical University
Tafila, 66110, Jordan

[2] Computer Engineering Department, Faculty of Engineering, Mutah University
Mutah, 61710, Jordan

**Abstract**
A new approach for signature region of interest pre-processing was presented. It used new auto cropping preparation on the basis of the image content, where the intensity value of pixel is the source of cropping. This approach provides both the possibility of improving the performance of security systems based on signature images, and also the ability to use only the region of interest of the used image to suit layout design of biometric systems. Underlying the approach is a novel segmentation method which identifies the exact region of foreground of signature for feature extraction usage. Evaluation results of this approach shows encouraging prospects by eliminating the need for false region isolating, reduces the time cost associated with signature false points detection, and addresses enhancement issues. A further contribution of this paper is an automated cropping stage in bio-secure based systems.

## 1. Introduction

There has been relatively little work done in the field of pre-processing of biometric images for the usage of biometric authentication and bio crypto key generation. Signature is one of behavioral biometric types. Signatures are a special case of handwriting subject to intrapersonal variation and interpersonal differences. This variability makes necessary to analyze signatures as complete images and not as collection of letters and words [1]. Any signature verification system built on five stages: data acquisition, pre-processing, feature extraction, comparison process and performance evaluation [2]. We concern in this paper on the pre-processing stage, where this stage consists of several steps: one of these steps is noise removing [3].

Another one which proposed in [4] is a noise filtering which is applied to remove the noise caused by the scanner, where the image is cropped to the bounding rectangle of the signature, then a transformation from color to grayscale and so forth to black and white, finally the bounding pixels
of the signature are marked and removed without eliminating the ending point until get the signature skeleton, i.e. Skeletonization, unless a pixel has less than three neighbors left, or any path between the neighbor pixels would be interrupted by the action. A non-uniformity correction for sensor elements, localization of the signature in the picture, extraction of the signature from the background, slicing, thresholding, segmentation, normalization and data reduction[5] are another steps for pre-processing. An experimental database has been acquired on a restricted size grid [6], Normalization includes basic techniques like, scaling, translation, and rotation etc [5] presented to avoid signature rotational problems. A given signature images threshold using Otsu's method [7]. A proposed approach consists of: image binarization, image normalization (resize), image thresholding, image enhancement using morphological operators, edge detection, and finally cropping. In signature verification technology, many algorithms used manual cropping such as in [2], while automatic cropping algorithms suffers from cropping information lost, since it is drawing rectangle rounded the signature then transfer the region inside it into another matrix (false rectangle). The region of interest term referred to object interest sub region of an image, leaving other regions unchanged. In signature verification system the signature it self is the ROI.

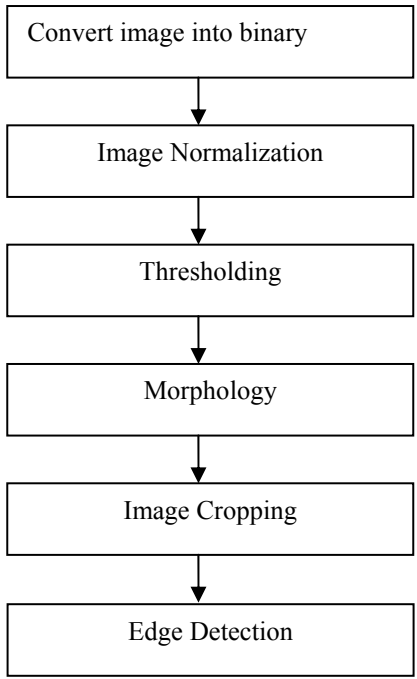

Fig. 1 Signature pre-processing stages

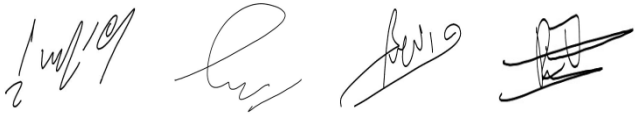

Fig. 1 Examples of signature

As in figure 2; the signature surrounded by unwanted area, this area may take role in post processing, such as increase the processing time. This area will be eliminated using auto cropping to keep only ROI for next stage of system verification or process generation of security based key.

## 2. Database

Approach has been performed and tested on base of 160 signatures. These signatures were acquired using pen tablet; the acquisitioned signatures were in JPEG format, 610X410, 24 colors.

## 3. Pre-Processing

Pre-processing is an important and diverse set of image preparation for next stage of image processing based application. Image pre-processing techniques have an important role in increasing the accuracy and performance of post processing application. Working directly on acquired images may degrade the final result of built on system; therefore we start with enhancement as pre-processing technique to sort out from low quality images as well to enhance the performance of the whole system.

### 3.1 Binarization

The first step in given approach is to convert the acquisitioned signature to binary form, i.e. black and white pixels. Working in this form is more useful than any other form, since it is easy to work with 2 bits representation of image.

| Coloured image (2D) | | | Binary image | | |
|---|---|---|---|---|---|
| 296 | 297 | 298 | 296 | 297 | 298 |
| 26 | 240 | 253 | 255 | 1 | 1 | 1 | 1 |
| 27 | 255 | 255 | 255 | 1 | 1 | 1 | 1 |
| 28 | 249 | 252 | 252 | 1 | 1 | 1 | 1 |
| 29 | 250 | 255 | 253 | 1 | 1 | 1 | 1 |
| 30 | 255 | 254 | 253 | 1 | 1 | 1 | 1 |
| 31 | 249 | 250 | 252 | 1 | 1 | 1 | 1 |
| 32 | 255 | 252 | 254 | 1 | 1 | 1 | 1 |
| 33 | 255 | 253 | 254 | 1 | 1 | 1 | 1 |
| 34 | 249 | 253 | 254 | 1 | 1 | 1 | 1 |

Fig. 2 colored Vs binary image matrixes.

The required time to process colored image is longer than binary one, as example applying radon transform on different types of image (colored Vs binary) shows the following deference's.

Table 1: processing time comparison

| Type | Coloured(3D) | Gray scale(2D) | Binary |
|---|---|---|---|
| **Time** | 6.2340 | 2.3440 | 2.0780 |

### 3.2 Image normalization (resizing).

Normalization is one of the most important steps of image pre-processing techniques. Experimental Database has been acquired on a restricted size grid, Normalization includes basic techniques like, scaling, translation, and rotation etc, is used to avoid scaling and rotational problems. As well in systems where the writer may have to sign on tablets or paper with different active areas, signature size normalization may be required. People normally scale their signatures to fit the area available for the signature. However size difference may be a problem in comparing the two signatures. Therefore signatures are normalized with respect to width, height or both [8]. To achieve logical results, the signatures must have the same size, which means normalized one, in our approach the reference sizes are [128 256].

### 3.3 Thresholding

We used Otsu's method or basically gray threshold computed used global image threshold values. Otsu's method is based on threshold selection by statistical criteria [7]. Otsu suggested minimizing the weighted sum of within-class variances of the object and background pixels to establish an optimum threshold. Threshold value based on this method will be between 0 and 1, after achieve this value we can segment an image based on it.

### 3.4 Morphological operators

The Mathematical Morphology is calculated based on simple mathematical concepts from set theory, morphological operators are useful for binary image, it views binary image as asset of its foreground (1-valued) pixels, and set operations such as union and intersection can be applied directly to binary image sets [9]. In our approach, we used bridge to connect discontinuity of pixels, then used remove operator to remove interior pixels keeping the boundaries of signature. A skeletonization is used to remove pixels on the boundaries of objects but does not allow objects to break apart. The pixels remaining make up the image skeleton.

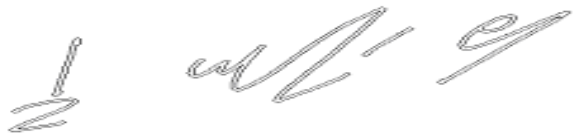

Fig. 3  bridge then remove operator's effect

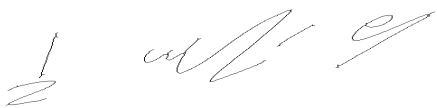

Fig. 4  skeleton operator effect

We did not use thinning because of this operator may cause lost of information, but using remove and bridge operator save the greatest amount of information, since they keep the boundary of signature.

### 3.5 Edge Detection

Edge detection is a fundamental tool used in most image processing applications to obtain information from the frames as a precursor step to feature extraction and object segmentation. This process detects outlines of an object and boundaries between objects and the background in the image [10]. Edge is a set of connected pixels that lie on the boundary between two regions. There are many operators could be used with edge detection technique such as canny, sobel, prewitt and roberts. In our proposed approach we used can for the reason of clearness, table 2 shows some of comparison results between these operators on our prepared database.

Table 2: edge detection operatores copmarsion, R is subject reader

| Operator | Clock | $R_1$ | $R_2$ | $R_3$ | $R_4$ | Clearance |
|---|---|---|---|---|---|---|
| Canny | 1.0320 | 97% | 96% | 97% | 96% | 96% |
| Sobel | 0.2500 | 70% | 66% | 66% | 69% | 67% |
| Prewitt | 0.0780 | 45% | 22% | 18% | 24% | 23% |
| Roberts | 0.0620 | 80% | 81% | 80% | 81% | 80% |

The comparison was decided subjectively by given edged figures to the group of subject's researchers who are working in the image processing field to decide the clearance according to the following factors:
- Contrast clearance.
- Edge clearance.
- Continuity clearance.

The canny method differs from the other edge-detection methods in that it uses two different thresholds (to detect strong and weak edges), and includes the weak edge in the output only if they are connected to strong edge. This method is therefore less likely than the other to be fooled by noise, and more likely to detect true weak edges.

## 4. Image Normalization (Auto Cropping)

In this stage the ROI is determined using auto cropping approach. Region of Interest (ROI) is the signature object itself. Using cropping we segment the signature smoothly. Signature cropping process is less complexity in process and time, since the area under process will be reduced. Applying Radon transform to check the time complexity approved this hypothesis, figure 6 show this result.

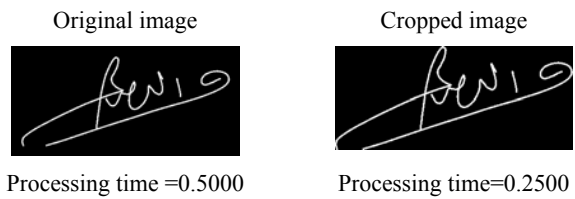

Original image        Cropped image

Processing time =0.5000     Processing time=0.2500

Fig. 6  Radon check of original and cropped images

Two types of cropping technique were used; manual and automatic cropping. Manual cropping is achieved using Matlab® function (imcrop), but it may cause false cropping rectangle and it is tedious work. While automatic cropping is saving more work and it is reducing a processing time over and above the cropping rectangle is truly detecting. Auto cropping approach, Figure 8, is firstly, determine the positions of ones in image, then calculate minimum and maximum coordinates from these positions, minimum coordinate will be the first corner (upper-left), the second one (lower-right) will be determined by subtract minimum coordinate from maximum coordinate. After that image cropping will be used with these corners as:

I2=imcrop(I1, x1 y1 x2 y2);
where;
I2: cropped image (output),
I1= original image (input),
x1,y1: first corner,
x2,y2: second corner.

According to the previous mechanism a good and true cropping rectangle was determined, so no lose in the information meaning of the signature object. Figure 7 shows this results, while Figure 8 show the whole flowchart of auto cropping approach.

Original image      Cropped image

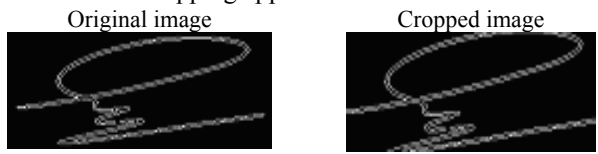

Fig. 5  Auto cropping results

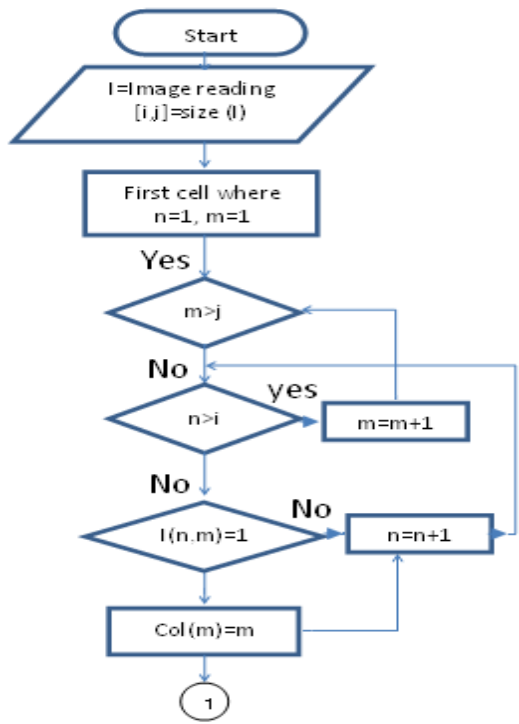

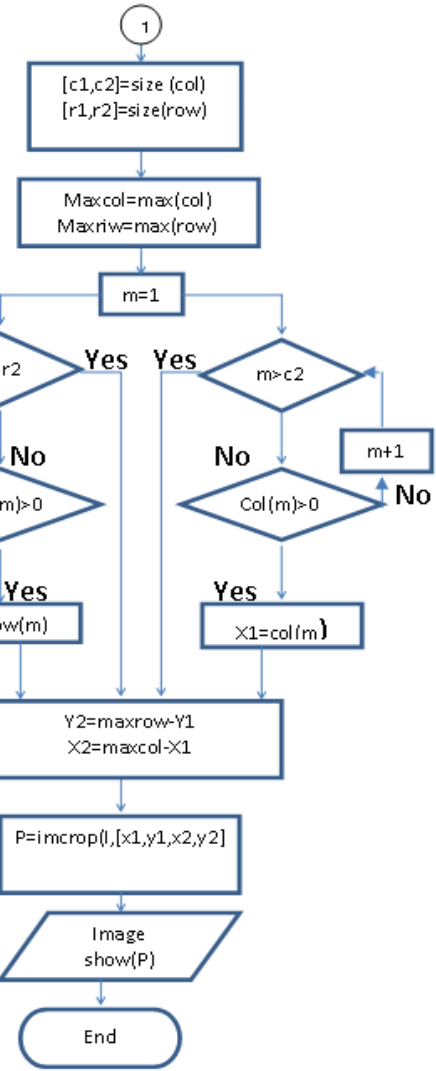

Fig. 6  auto cropping flowchart

## 5. Conclusion

This paper proposed new approaches of pre-processing and auto cropping of signature images, where the given approach shows high accuracy of signature object detection and segmentation, the performance of this approach was shown through its speed time, and keeping the content information of the signature object without losing.

## References

[1] Fang, B. Off-line signature verification with generated training samples. in IEE Proc.Vis. Image and Signal Processing. 2002.

[2] Plamondon, R. and G. Lorette, Automatic signature verification and writer identification: The state of the art. Pattern Recognition, 1989 22(2): p. 107-131.

[3] Abuhaiba, I.S., Offline Signature Verification Using Graph Matching. Turk. J. Elec. Engin, 2007. 15(1): p. 89-104.

[4] Kővári, B. Extraction of Dynamic Features for Off-line Signature Analysis. in Automation and Applied Computer Science Workshop (AACS). 2007.

[5] Santosh, K. and N. Cholwich, A Comprehensive Survey on On-Lie Handwriting Recognition Technology and its Real Application to the Nepalese Natural Handwriting. Kathmandu University Journal of Science, Engineering and Technology, 2009. 5(1): p. 31-55.

[6] Fierrez-Aguilar, J., Alonso-Hermira, N., Moreno-Marquez, G., Ortega-Garcia, J., An off-line signature verification system based on fusion of local and global information. Proc. of BIOAW, Springer LNCS-3087, 2004: p. 295–306.

[7] Otsu, N., A threshold selection method from gray level histograms. IEEE Trans. Systems, Man and Cybernetics, 1979. 9: p. 62-66.

[8] Muralidharan, N. Signature Verification: A Popular Biometric technology. in Second LACCEI International Latin American and Caribbean Conference for Engineering and Technology. 2004.

[9] Gonzalez, R.C., R.E. Woods, and S.L. Eddins, Digital Image Processing USING MATLAB. 2004: Pearson Prentice Hall.

[10] Neoh., H.S. and A. Hazanchuk. Adaptive Edge Detection for Real-Time Video Processing using FPGAs. in GSPx 2004 Conference. 2004.

**Bassam AL-Mahadeen** is an assistant professor at Math & IT department, faculty of science at Tafila Technical University, Jordan. He obtained his PhD in CIS from Arab Academy for Banking & Financial Sciences, Jordan in 2005. He also holds MSc in Computer Science from Al Al-Byte University, Jordan in 1999, and BSc in Computer Science from Mutah University, Jordan in 1992. His research interests include Wireless image sensor networks, RFID applications, Healthcare Simulation and Modeling, image processing, and Interconnection Networks. He was the Director of Computer Center at Tafila Technical University, and Full Time Lecturer at Mutah University. He is a member of IAJIT.

**Mokhled S. AlTarawneh** is an assistant professor at computer engineering department, faculty of engineering at Mutah University, Jordan. He is also a director of computer center, Mutah University. His research interests include computer vision, pattern recognition, image processing, image quality assessment, and biometrics. Altarawneh is a member of Jordan Engineers Association (JEA), member of Arab Computer Society (ACS), member of IET, UK and International Engineers Association (IAENG). AlTarawneh holds a Doctorate degree in computer engineering from Newcastle University, United Kingdom. He received the master's degree from University of the Ryukyus, Okinawa, Japan, the BE degree in computer engineering from Azerbaijan Technical University, Baku, Azerbaijan.